\RequirePackage{fix-cm}
\documentclass[smallextended]{svjour3}       
\smartqed  
\usepackage{graphicx}
\usepackage{xcolor}
\begin{document}

\title{Deep Learning Based Framework for Iranian License Plate Detection and Recognition
}


\author{Mojtaba Shahidi Zandi         \and
        Roozbeh Rajabi 
}


\institute{R. Rajabi \at
              Communications and Electronics Department, Faculty of Electrical and Computer Engineering, Qom University of Technology, Qom, Iran
              \email{rajabi@qut.ac.ir}           
}

\date{Received: date / Accepted: date}

\maketitle

\begin{abstract}
License plate recognition systems have a very important role in many applications such as toll management, parking control, and traffic management. In this paper, a framework of deep convolutional neural networks is proposed for Iranian license plate recognition. The first CNN is the YOLOv3 network that detects the Iranian license plate in the input image while the second CNN is a Faster R-CNN that recognizes and classifies the characters in the detected license plate. A dataset of Iranian license plates consisting of ill-conditioned images also developed in this paper. The YOLOv3 network achieved 99.6\% mAP, 98.26\% recall, 98.08\% accuracy, and average detection speed is only 23ms. Also, the Faster R-CNN network trained and tested on the developed dataset and achieved 98.97\% recall, 99.9\% precision, and 98.8\% accuracy. The proposed system can recognize the license plate in challenging situations like unwanted data on the license plate. Comparing this system with other Iranian license plate recognition systems shows that it is Faster, more accurate and also this system can work in an open environment.
\end{abstract}

\section{Introduction}
\label{section,Indroduction}
Automatic License Plate Recognition (ALPR) systems identify vehicle license plates automatically and reduce human resources; so in the recent years has become more important. There are more than 1.2 billion vehicles in the world and over 19 million of these numbers are belong to Iran and it is enough reason for importance of developing more reliable ALPR systems in this country. ALPR systems play important role in smart cities such as parking lots, traffic surveillance, access control, toll payments, and so on \cite{ref.plate}.

Traditional systems employ features that has been regarded by humans to represent the features of the images. Traditional systems require human-designed models for recognition responses from raw input pixels. But systems based on deep learning techniques automatically select the features themselves and do not need humans to design features \cite{ref.CNN1}. These systems learn low-level representations of  the underlying data by modifying ﬁlters \cite{ref.deeplearning}. Deep learning has several methods and most common method in image processing field is Convolutional Neural Networks (CNNs) \cite{IET_IP_19_CNN}. CNNs are based on sharing weights and sparse connections that giving high learning potential \cite{ref.endtoendCNN}. Two biggest challenges in CNNs there are that first of them is demand for large amounts of training samples and second one high computational cost \cite{ref.deeplearningForecasting}. In recent years, CNNs have become immensely more powerful as have been discovered clever solutions to reduce the amount of training samples and computational cost \cite{ref.SingleResidential}. Since the CNNs in many task within the field of visual recognition are state-of-the-art it is believed that also are state-of-the-art within ALPR \cite{SpringerMultimedia_Lu2019}.

The ALPR systems have a complex task in changing condition and with many variations \cite{ref.Bayesian}. The camera maybe have different distance and angle with vehicle, the resolution of cameras are different, uneven lighting, reflection and refraction of light, big background interference, large angle incline, dust and moisture are other complexities. A good ALPR system in addition to working good in mentioned conditions, should also detect located plates in different parts of the image, and if there are more than one plate in the image, the system should locate all of them. Many of the systems claim accuracy more than 90\% \cite{SpringerMultimedia_Asif2019}. However, in most cases claimed accuracy is incomparable to other systems  because in each one systems different data sets are used and there is not an accepted universal data set. The complexity of the environment in recognizing license plates in different test sets will significantly impact the accuracy and makes direct comparisons of the accuracy without considering these complexities meaningless \cite{SpringerMultimedia_Islam2020}. Traditional algorithms often work well under certain circumstances but in open environment like uneven lighting, reflection and refraction of light, big background interference, large angle incline, dust and moisture fail to achieve satisfactory accuracy \cite{ref.CNN-RNN}.

ALPR systems often have tree main steps: license plate detection, license plate segmentation and license plate recognition \cite{IET_IP_18_SVM}. The license plate detection step get the input image then locate plates in the image. For this step five main approaches are proposed: 1- Edge-based approaches: these approaches detect license plate by knowing that the license plate is rectangular with a known aspect ratio \cite{ref.morphological,ref.template-matching}. 2- Color-based approaches: these approaches detect license plates by exploiting the fact that license plates often has a different color compared to the background \cite{ref.colorfeatures}. 3- Texture-based approaches: these approaches detect license plates by use the unconventional pixel density distribution inside the license plate \cite{ref.detectin-texture}. 4- Character-based approaches: theses approaches detect plate as strings of characters \cite{ref.detection-character}. 5- Learning-based approaches: these approaches use machine learning techniques to recognition characters based on features and now this approaches usually use CNNs \cite{ref.2YOLOv2,ref.CRNN-12,ref.2cnn}. The license plate segmentation step segment the character of the license plate that located with previous step. For this step two main approaches are proposed: 1- Projection-based approaches: these approaches segmentation characters by knowing that characters and the background of the license plate have different colors that giving opposite values after binarization of the image \cite{ref.templatematching1}. 2- Pixel-connectivity-based approaches: these approaches segmentation characters by  labeling all connected pixels in the binary image. The pixels having the same properties as the characters are considered part of the character \cite{ref.colorfeatures}. The license plate recognition step recognize characters and for this step two main approaches are proposed: 1- Template matching approaches: these approaches recognition carachters by comparing the similarity of the character and a template \cite{ref.templatematching1,ref.templatematching2}. 2- Learning-based approaches: these approaches use machine learning techniques to recognition characters based on features and now this approaches usually use CNNs \cite{ref.2YOLOv2,ref.CRNN-12,ref.2cnn,ref.AlexNet}.

\begin{figure}[t!]
	\centering
	\includegraphics[width=1\textwidth]{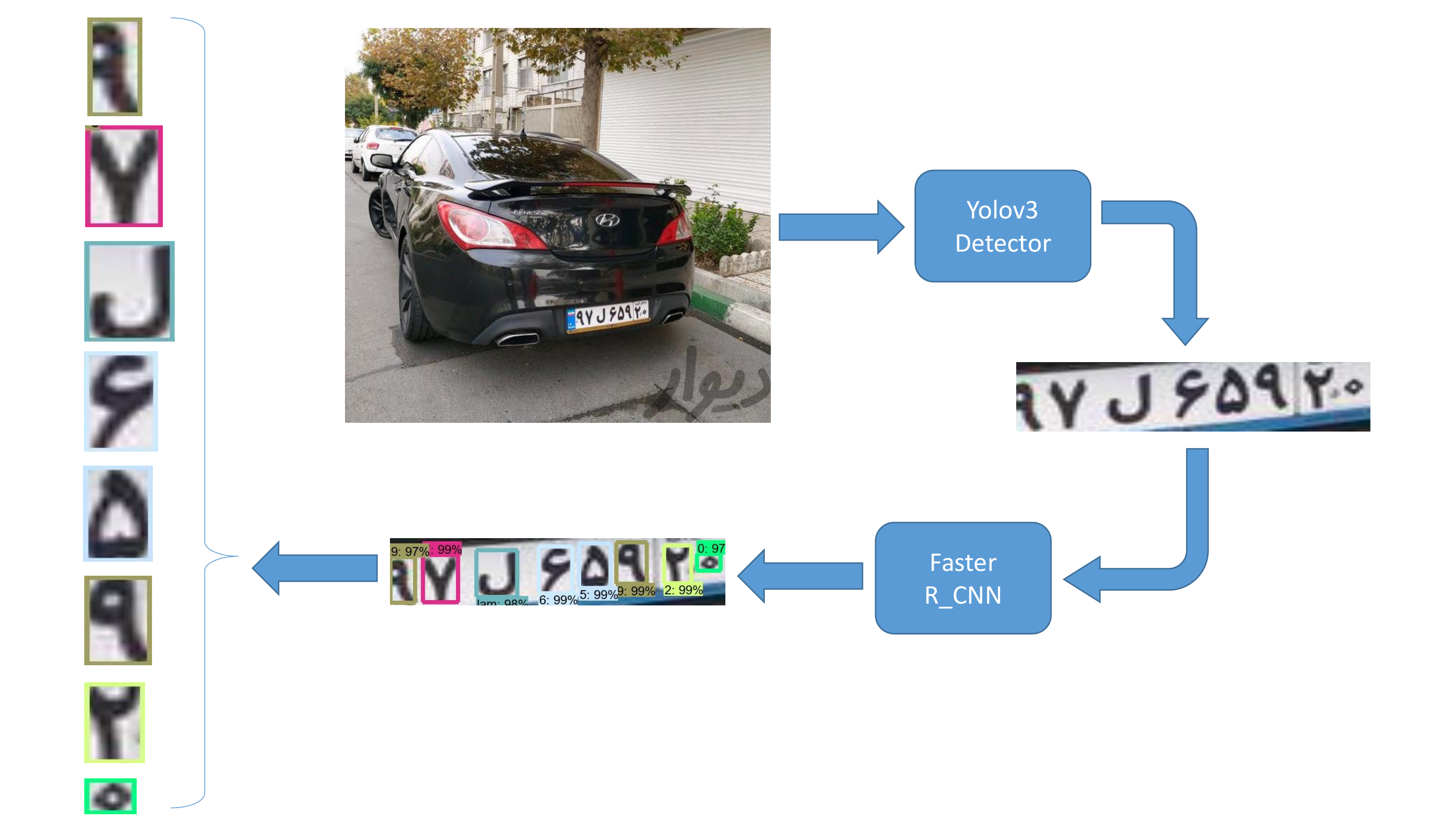}
	\caption{Architecture of the proposed license plate Recognition system} 
	\label {figure.ALPRS}
\end{figure}

In real world applications, the data acquisition conditions are changing and traditional machine learning methods may fail in recognizing the license plates accurately. So in this paper we proposed a framework based on deep learning schemes to overcome these limitations. In this paper, Iranian License Plate Recognition System based on deep convolutional neural networks are proposed. This system consists of two main part. First part is license plate detection based on YOLOv3 \cite{ref.YOLOv3} network. In this part the image of Iranian car feed into the system and the YOLOv3 detect plate or plates from the image and crop them as output. Second part is license plate recognition based on the Faster R-CNN \cite{ref.FasterR-CNN,IET_IP_20_FasterRCNN}. In this part the license plate that have been cropped from YOLOv3 are coming to Faster R-CNN and all numbers or alphabets in the plate are detected and recognition. The architecture of proposed system are shown in Fig. \ref {figure.ALPRS}.

The following steps have been taken to prepare the proposed system: 
1- preparing and labeling dataset for YOLOv3 (building YOLOv3 dataset) 2- training YOLOv3 network 3- testing YOLOv3 network 4- preparing and labeling images for Faster R-CNN (building Faster R-CNN dataset) 5- training Faster R-CNN network 6- testing Faster R-CNN network

Since there is not any Iranian license plate dataset, we developed a dataset for Iranian license plate recognition in this paper. The images crawled on the web and the dataset consists of car images in different ill-conditioned situations. Total number of images in the dataset is 2316.

The rest of the paper is organized as follows: section \ref{section.YOLOv3} explains the architecture of the proposed YOLOv3 for detection of license plates. Section \ref{section.FasterR-CNN} explains the architecture of the proposed Faster R-CNN for recognition of alphabets and numbers. Section \ref{section.dataset} explains the gathered dataset for testing and training of these two networks. In section \ref{section.Experimentsandresults} experiments and results are explained and the proposed system has been compared with other systems and section \ref{section.conclusion} concludes the paper.

\section{License plate detection using YOLOv3}
\label{section.YOLOv3}
YOLO (2015) \cite{ref.YOLO} was a CNN with one-shot architecture. The one-shot architectures processes the image with a single CNN without the need of region and object proposals so this architectures  is simpler than R-CNN. YOLO mAP score  on VOC 2007 dataset was 63.4  with a processing time of 20 ms per image. YOLOv2 (2016) \cite{ref.YOLOv2} is an improved version of the YOLO network. YOLOv2 training with a multi-scale method, so it can run in different sizes and offer a trade-off between speed and accuracy \cite{IET_IP_19_YoloL}. The version that runs at 67 FPS achieves a mAP on VOC 2007, while a slower version that runs at 40 FPS achieves a 78.6 mAP. YOLOv3 (2017) \cite{ref.YOLOv3} is the third version of YOLO \cite{ref.slidingYOLO} network. YOLOv3 uses a few tricks to improve training and increase performance like multi-scale predictions (prediction of small, medium amd large objects in image) and use a better backbone classifier (YOLOv3 uses Darknet-53 while YOLOv2 uses Darknet-19). YOLOv3-320 gives a MAP of 51.5 with 45 FPS on COCO dataset for IOU 0.5 and YOLOv3-608 gives a MAP of 57.9 with 20 FPS on COCO dataset for IOU 0.5. 

\subsection{Darknet-53}
YOLOv3 uses Darknet-53 for performing feature extraction \cite{ref.YOLOv3} and 53 means this network has 53 convolutional layers. This new network is much more powerful than Darknet-19 (used in YOLOv2). Darknet-53 has similar performance to ResNet-152 but it is faster. Darknet-53 architecture is shown in Table \ref{table.Darknet53}.

\begin{table}[!t]
	\centering
	\renewcommand{\arraystretch}{1.3} 
	\caption{Darknet-53}
	\label{table.Darknet53}
	\begin{tabular}{|p{0.8cm}|p{2.2cm} p{0.9cm} p{1.5cm} p{1.2cm}|}
		\hline
		Repeat&Type &Filters & Size & Output \\
		\hline
		&Convolutional & 32 &  3\(\times\)3 &  256\(\times\)256 \\
		&Convolutional & 64 &  3\(\times\)3 / 2 &  128\(\times\)128 \\
		\hline
		&Convolutional & 32 &  1\(\times\)1 &  \\
		1\(\times\)&Convolutional & 64 &  3\(\times\)3 &   \\
		&Residual& & &128\(\times\)128 \\
		\hline
		&Convolutional & 128 &  3\(\times\)3 / 2 &  64\(\times\)64 \\
		\hline
		&Convolutional & 64 &  1\(\times\)1 &  \\
		2\(\times\)&Convolutional & 128 &  3\(\times\)3 &   \\
		&Residual& & &64\(\times\)64 \\
		\hline
		&Convolutional & 256 &  3\(\times\)3 / 2 &  32\(\times\)32 \\
		\hline
		&Convolutional & 128 &  1\(\times\)1 &  \\
		8\(\times\)&Convolutional & 256 &  3\(\times\)3 &   \\
		&Residual& & &32\(\times\)32 \\
		\hline
		&Convolutional & 512 &  3\(\times\)3 / 2 &  16\(\times\)16 \\
		\hline
		&Convolutional & 256 &  1\(\times\)1 &  \\
		8\(\times\)&Convolutional & 512 &  3\(\times\)3 &   \\
		&Residual& & &16\(\times\)16 \\
		\hline
		&Convolutional & 1024 &  3\(\times\)3 / 2 &  8\(\times\)8 \\
		\hline
		&Convolutional & 512 &  1\(\times\)1 &  \\
		4\(\times\)&Convolutional & 1024 &  3\(\times\)3 &   \\
		&Residual& & &8\(\times\)8 \\
		\hline
		&Avgpool& &Global& \\
		&Connected& &1000&\\
		&Softmax& & &\\
		\hline
	\end{tabular}
\end{table}

\subsection{YOLOv3 architecture}
\begin{figure*}[t!]
	\centering
	\includegraphics[width=1\textwidth,height=10cm]{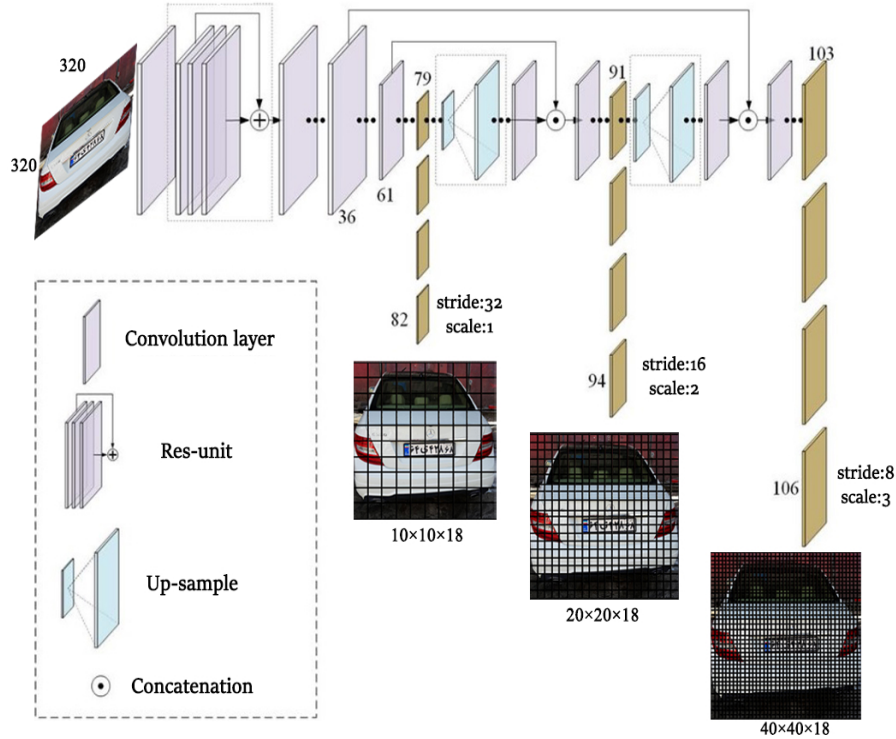}
	\caption{Architecture of proposed YOLOv3 network. The input image has 320\(\times\)320 pixels and YOLOv3 network detect license plate on input image with 3 scales in stride 32,16,8 for large, medium and small license plate respectively. }
	\label{figure.YOLOv3}
\end{figure*}

YOLOv3 \cite{ref.YOLOv3} is the third version of YOLO (You Only Lock Once) network. YOLOv3 use Darknet-53 for performing feature extraction. The Darknet-53 has 53 convolutional layers and in detection stage the YOLOv3 has 106 layers. The architecture of YOLOv3 have been show in Fig. \ref{figure.YOLOv3}.

\begin{figure}[!t]
	\centering
	\includegraphics[width=8cm,height=8.91cm]{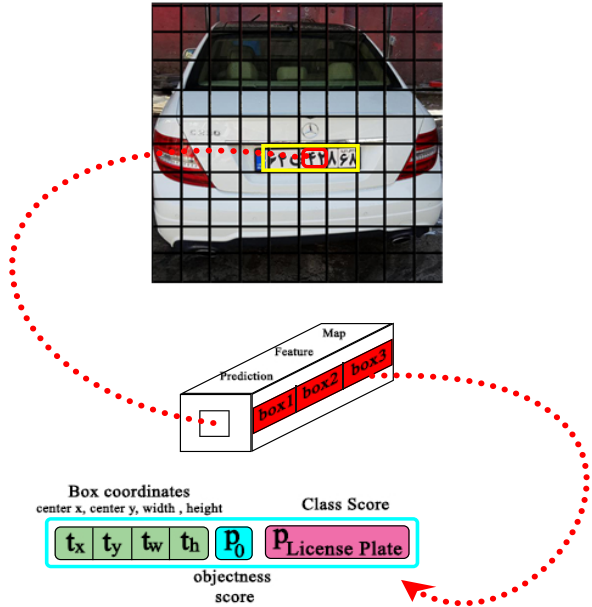}
	\caption{ Architecture of a grid cell in feature map. \(t_{x}\), \(t_{y}\), \(t_{w}\), \(t_{h}\) are box coordinates that are center x , center y, box width and box height respectively. \(P_{0}\) show probability that box are background and \(P_{License-plate}\) show probability that box are consist of a license plate.}
	\label{figure.YOLOv3kernel}
\end{figure}

The most salient feature of YOLOv3 compared with previous YOLO is that it makes detection at three different scales which down sampling the input image by 32, 16 and 8 respectively \cite{ref.YOLOv3}. In YOLOv3 its eventual output is generated by applying a 1 \(\times\) 1 kernel on feature map of three different size at three different places in network. The shape of detection kernels come from  Eq.\ref{equation.YOLOv3kernel}. 

\begin{equation}
1 \times 1 \times (B \times (5 + C)) 
\label{equation.YOLOv3kernel}
\end{equation}
where B is the number of bounding box that a cell on feature map can predict and C is the number of classes. In the proposed YOLOv3 just one class named License Plate is considered and B = 3 so the detection kernel size become 1\(\times\)1\(\times\)18 [1\(\times\)1\(\times\) (3 \(\times\) (5 + 1))]. A grid cell architecture in feature map is shown in Fig. \ref{figure.YOLOv3kernel}. Since in proposed LPDS the input image is 320 \(\times\) 320, the detection give feature maps of 10\(\times\)10\(\times\)18, 20\(\times\)20\(\times\)18 and 40\(\times\)40\(\times\)18. The 10\(\times\)10 layers responsible of detection large objects, whereas 20\(\times\)20 layers detect medium objects and 40\(\times\)40 layers detect small objects. This LPDS use 9 anchors; three biggest anchors for large scale, next three for medium scale and the last three for the small scale. YOLOv3 at each scale can predict 3 boxes using 3 anchors, so this LPDS predict 6300 boxes in total.

\section{License Plate Recognition with Faster R-CNN}
\label{section.FasterR-CNN}
After cropping the license plate from the input image, numbers and alphabets in the plate should be recognize separately. This is accomplished using Faster R-CNN \cite{ref.FasterR-CNN}. 
In the R-CNN family; after R-CNN and Fast R-CNN, the Faster R-CNN have been introduced. Region-based CNN (2014) \cite{ref.R-CNN} achieved a mAP score of 66 with a processing time of 20 second per image in VOC 2007 dataset. Fast R-CNN (2015) \cite{ref.FastR-CNN} is an improved and simplified version of the R-CNN model that the speed and accuracy have been increased. Fast R-CNN achieved a mAP score of 70.0 on the VOC 2007 with a processing time of 2 seconds per image, 10 times faster than R-CNN. Faster R-CNN (2016) \cite{ref.FasterR-CNN} is  improvement version of Fast R-CNN by changing the region proposer completely that results speed and accuracy improved. Faster R-CNN achieved a mAP score of 73.2 with a processing time of 140 ms per image. This is over 10 times faster than Fast R-CNN and over 100 times faster than R-CNN. In fainal by changing the underlying architecture from VGGnet to the ResNet mAP score of 76.4 is achieved.

\begin{figure*}[h!]
	\centering
	\includegraphics[width=\textwidth,height=8.91cm]{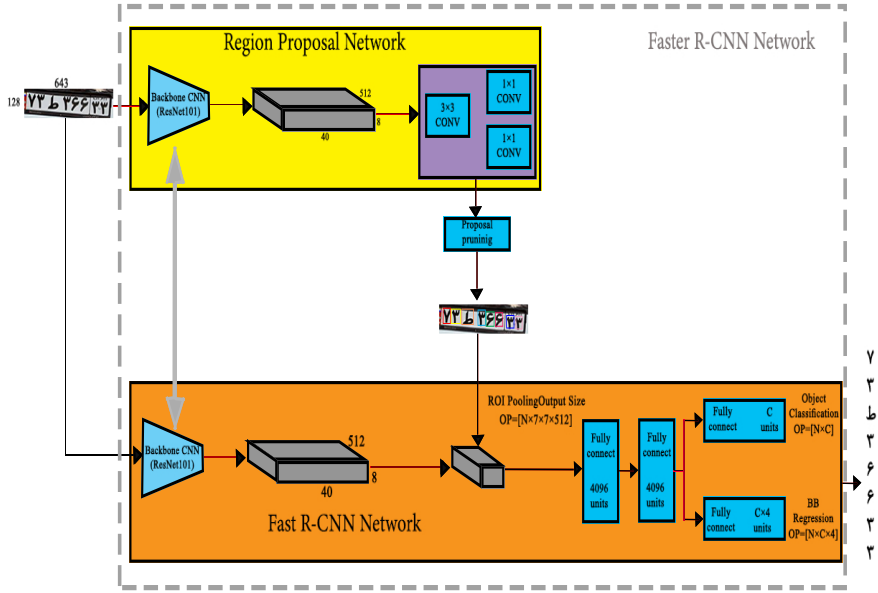}
	\caption{ Faster R-CNN architecture. The Faster R-CNN architecture consists of RPN as a region proposal algorithm and the Fast R-CNN as a detector network. In the proposed Faster R-CNN a ResNet101 are used as a feature extractor.}
	\label{figure.Faster R-CNN}
\end{figure*}

The Faster R-CNN architecture consist of RPN as a region proposal algorithm and the Fast R-CNN as a detector network. In proposed Faster R-CNN a ResNet101 \cite{ref.ResNet} are used as Feature extractor. Faster R-CNN architecture has been shown in Fig. \ref{figure.Faster R-CNN}. 

\subsection{ResNet}
\begin{figure*}[h!]
	\centering
	\includegraphics[width=1\textwidth,height=7cm]{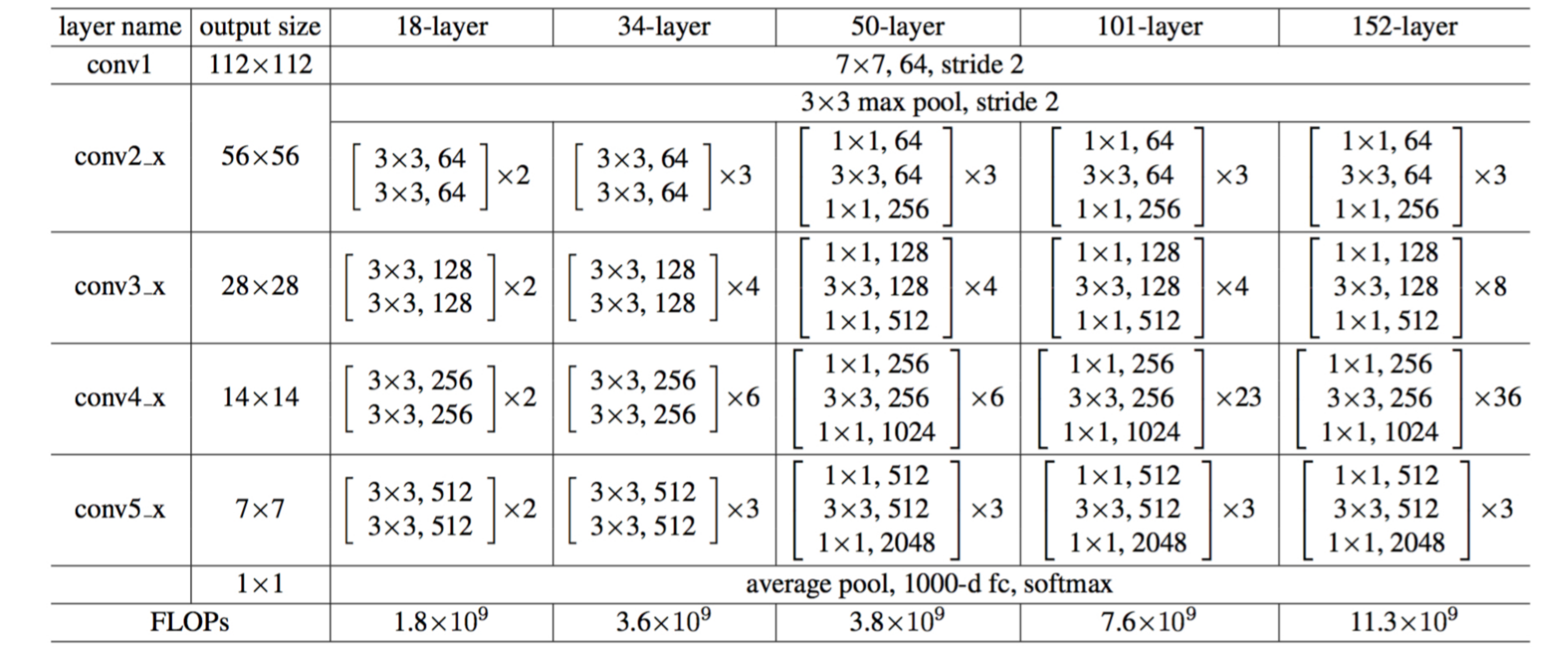}
	\caption{ ResNet family architecture \cite{ref.ResNet}. In proposed system ResNet-101 that is shown in column 3 used.}
	\label{figure.ResNet}
\end{figure*}

ResNet \cite{ref.ResNet} won the ILSVRC 2015 and represents a new revolutionizing way of building CNNs, called residual CNNs. This was achieved by using skip connections. A skip connection is a connection used by the input signal to bypass a number of layers. With skip connection technique, CNNs with over 1000 layers are trainable. In the field of general object classiﬁcation residual CNNs remain the state-of-the-art (June 2017) as in the ILSVRC 2016 no revolutionary architectures were submitted. ResNet family architecture is shown in Fig. \ref{figure.ResNet} \cite{ref.ResNet}. 

\subsection{RPN (Region Proposal Network)}
Just before feeding into RPN, image is resized to size of 128 by 640 pixels. First in the RPN, input image being fed into ResNet-101 (backbone convolutional neural network). As the stride in proposed system are 16 the output feature of backbone network  is a feature map with 8\(\times\)40\(\times\)512 size (this means that two consecutive pixels in the backbone (ResNet-101) output features correspond to 16 pixels in the input image). The Network, for every point in the feature map should learn whether an object is present in the input image at its corresponding location and estimate its size. This is done by placing a set of anchors on input image for each location on the output feature map for the backbone network. This anchors indicate possible objects in various size and aspects ratios at this location. In the proposed system the anchors uses 4 scale of box area and 3 aspect ratio so for every location in feature map 12 (4\(\times\)3) location in input image is being correspond. So the output feature map corresponding 3840 (8\(\times\)40\(\times\)12) anchors in total. 

In RPN and after backbone network, there is a convolutional layer with 3\(\times\)3 filters, 1 padding and 512 output channels. The output channels is connected to two 1\(\times\)1 convolutional layers for classification and box-regression (this classification just determine that the box is object or not) \cite{ref.FasterR-CNN}.

\subsection{Fast R-CNN (as a detector for Faster R-CNN)} 
The Fast R-CNN detector also consist a CNN backbone (ResNet101), an ROI pooling layer and fully connected layers that followed by two branches for classification and bounding box regression. 
In this section first the input image passed through backbone CNN (ResNet101) for feature map extraction of size 8\(\times\)40\(\times\)512 (this is same as extracting feature map in RPN and just weight sharing between RPN backbone and Fast R-CNN backbone).
In the next stage, the bonding box proposals from the RPN are used to pool features from the backbone feature map. This is done by the region of interest (ROI) pooling layer. The ROI pooling layer works by: a) Taking the region corresponding to a proposal from the backbone feature map. b) Dividing the region into a fixed number of sub-windows. c) Performing max-pooling to over this sub-windows to give fixed size output. 
The output of ROI pooling layer is N\(\times\)7\(\times\)7\(\times\)512 where N is the number of proposals from the region proposal algorithm. 
In the final step features pass through two fully connected layers, and then fed into sibling classification and regression branches. The classification layer has C unit for each of the classes (in this system C is equal to 25). The features pass through a softmax layer to get the classification scores. And the regression layer coefficient are used to improve the predicted bounding box. That is, all the classes have individual regressors with 4 parameters each corresponding to C\(\times\)4 output units in the regression layers \cite{ref.FasterR-CNN}.


\section{Datasets} \label{section.dataset}
After choosing and setting system architecture, to adjust the weights and reducing classification error, the systems should be trained \cite{SpringerMultimedia_Chowdhury2020}. This is done using transfer learning method that retrains the pretrained CNN with the new dataset without changing the architecture or reinitializing the weights \cite{IET_IP_19_Transfer,IET_IP_20_TransferHand}. As in the proposed system there are two networks, for training this system two datasets have been created; first training dataset for training YOLOv3 network and second one for training Faster R-CNN network. After training these two networks the system are ready and its performance in challenging situation should be tested. This is done by two other dataset ; first testing dataset for testing YOLOv3 network and second testing dataset for testing Faster R-CNN network. All images are gathered from the public access photos of the internet and databases will be publicly available for academic research after publishment of the paper.

\subsection{YOLOv3 Dataset}

\subsubsection{YOLOv3 Training dataset}
\begin{figure}[h!]
	\centering
	\includegraphics[width=1\textwidth]{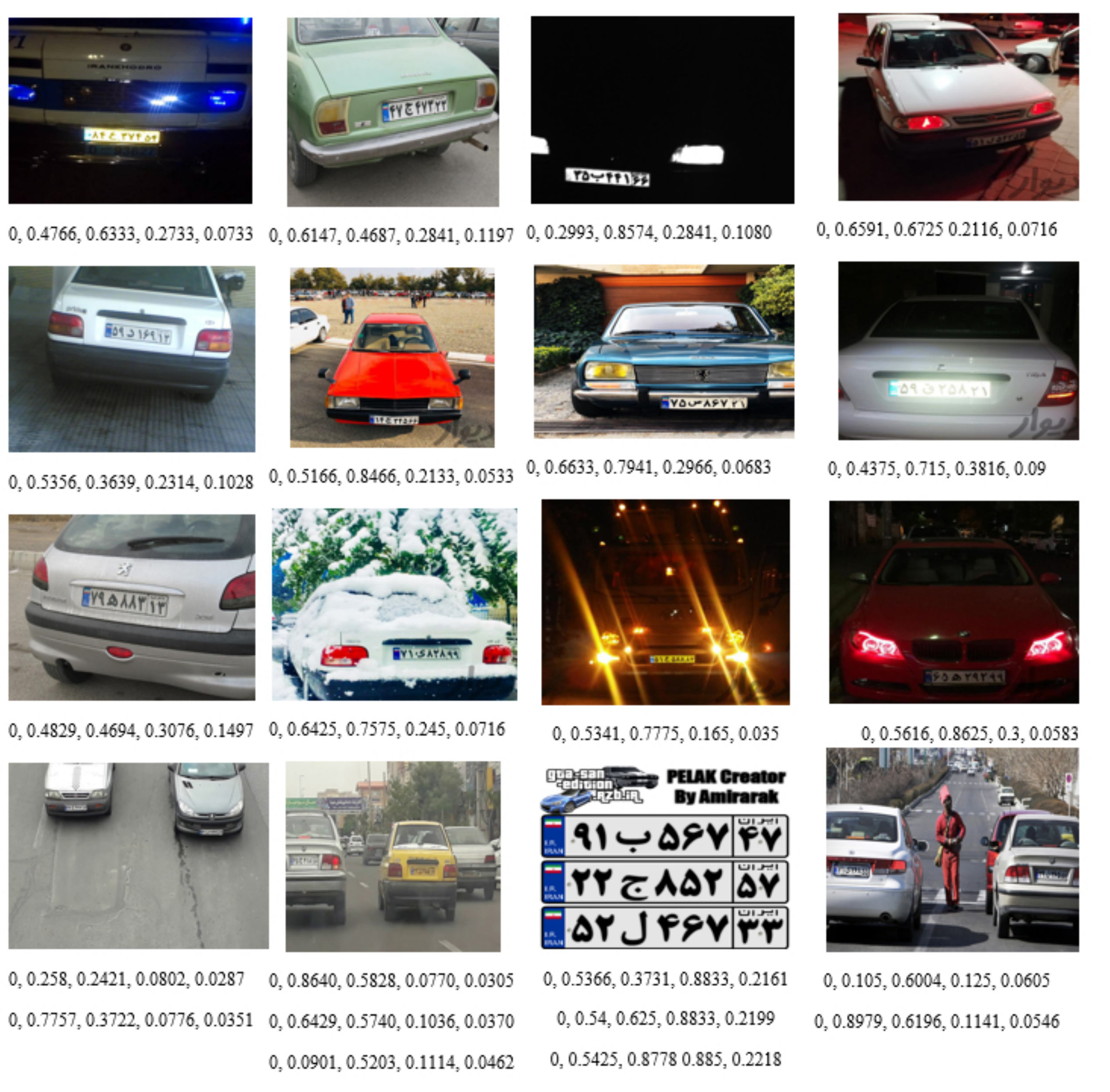}
	\caption{Some images of YOLOv3 training dataset and their labels in below of each images, that each row of label, point to a license plate in a picture. The first number points to its class label that zero means license plate. The next four numbers point to normalized center x, center y, width and height respectively.}
	\label{figure.YOLOv3Dataset}
\end{figure}

This dataset used for training YOLOv3 network and has very important role in overall system accuracy. Choosing its images in open environment, trains the system for work in challenging situation. So prepared dataset consists of 2316 Iranian car images and this images captured from Iranian car in open environment like in reflection and refraction of light, big background interference, large angle incline, dust, moisture, day, night, snowy and rainy weather, two or more cars in the pictures and so on. These images have been labeled and finally used for training the YOLOv3. This dataset is for detection and cropping plate from Iranian car image with YOLOv3 and just have one output class named License Plate. In Fig. \ref{figure.YOLOv3Dataset} some of the images used for training YOLOv3 with their labels have been shown. Each row of label, point to a license plate in a picture. The first number points to the output class that zero means license plate. The next four numbers point to normalized center x, center y, width and height respectively.

\subsubsection{YOLOv3 testing dataset}
This dataset is for testing the trained YOLOv3 network and to calculate its performance. Easily with a simple dataset can reach to a very high accuracy but for an exact accuracy this dataset also should consists of images in challenging environment. So a dataset consists of 512 Iranian car images for testing YOLOv3 network and calculating its performance prepared.

\subsection{Faster R-CNN Dataset }

\subsubsection{Faster R-CNN training dataset}

\begin{figure}[h!]
	\centering
	\includegraphics[width=1\textwidth]{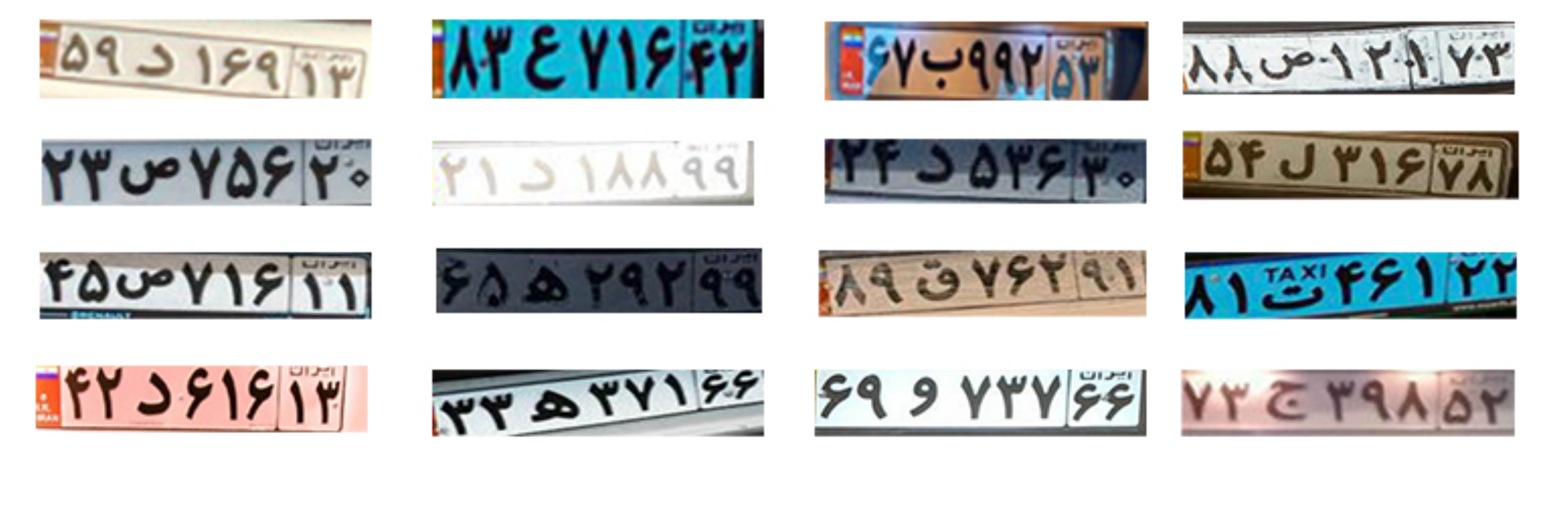}
	\caption{Some images of Faster R-CNN training dataset.}
	\label{figure.FasterRCNNDataset}
\end{figure}

The second dataset has been prepared for training Faster R-CNN network. For working system in challenging situations its images should be in challenging environment conditions. So this dataset consists of 1643 Iranian license plate images that captured from proposed YOLOv3 system. As every Iranian license plate consist of 7 numbers and 1 alphabet in total, for training Faster R-CNN, 13144 objects have been labeled. This dataset consist of 25 classes, 10 classes for numbers and 15 classes for alphabets. In Fig. \ref{figure.FasterRCNNDataset} some plates that have been labeled for training Faster R-CNN network is demonstrated.

\subsubsection{Faster R-CNN testing dataset}

This dataset is used for testing the Faster R-CNN network and calculating its performance. Also this dataset should be in challenging environment situations. The dataset consists of 517 license plates images and 4136 objects.

\section{Experiments and results} \label{section.Experimentsandresults}

\begin{figure*}[t!]
	\centering
	\includegraphics[width=1\textwidth]{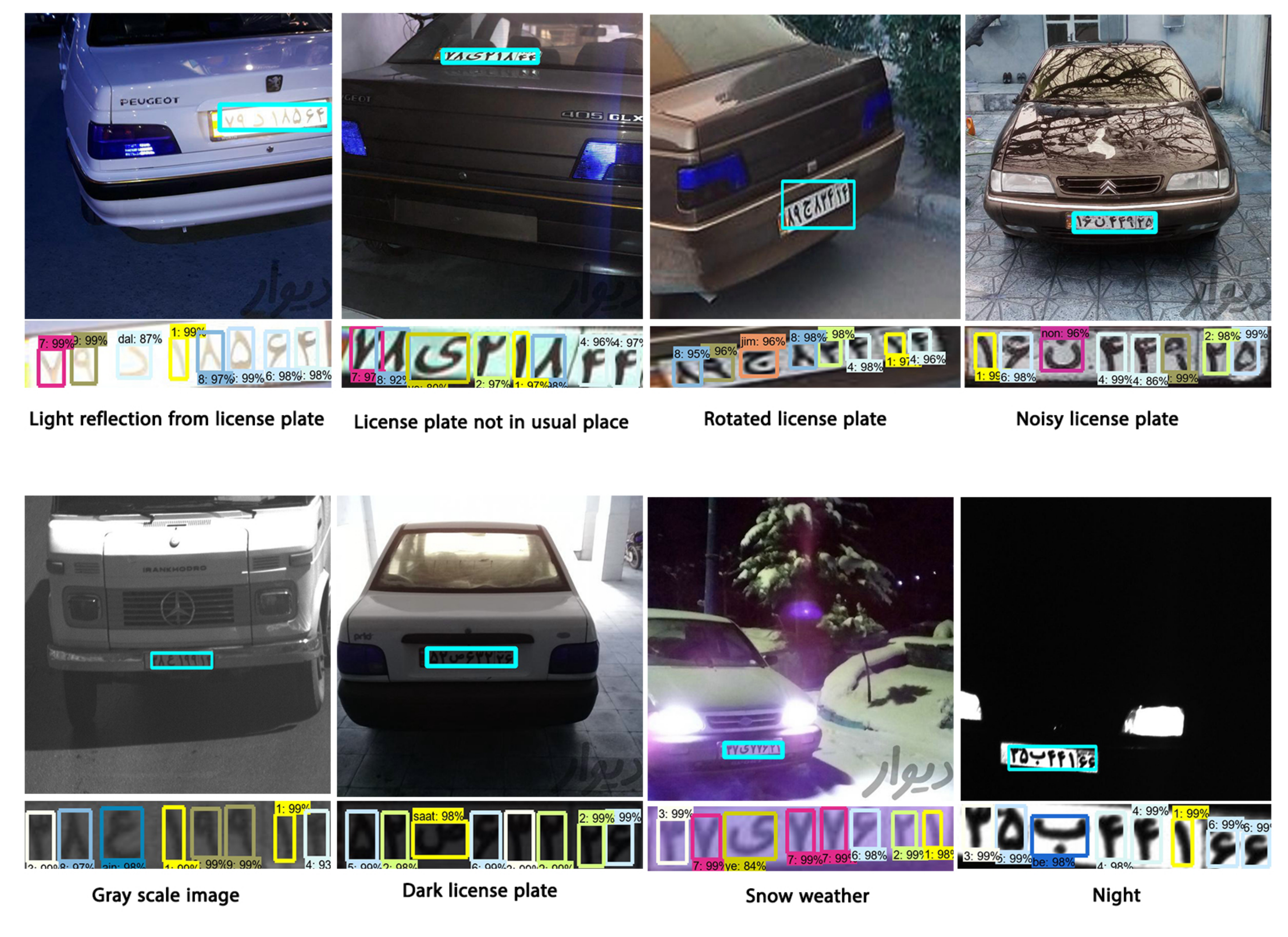}
	\caption{Examples of license plate detection and character detection and recognition using the proposed system in challenging conditions. }
	\label{figure.samples}
\end{figure*}

Now the systems are trained with training datasets and their performance calculated by running system on test datasets. Precision, recall and accuracy are the three main criteria for calculation of network performance. These criteria are calculated with equations \ref{equation.Precision}, \ref{equation.Recall} and \ref{equation.Accuracy} respectively. 

\begin{equation}
\centering
Precision =\frac{TP}{TP+FP}=\frac{TP}{All Detection}
\label{equation.Precision}
\end{equation}

\begin{equation}
\centering
Recall = \frac{TP}{TP+FN}=\frac{TP}{All Ground Truths}
\label{equation.Recall}
\end{equation}

\begin{equation}
\centering
Accuracy = \frac{TP+TN}{TP+TN+FP+FN}
\label{equation.Accuracy}
\end{equation}

where TP means true positive or correct detection, FP means false positive or wrong detection and FN means false negative or ground truth not detected and finally the TN means that the result should be negative and is negative. 
In Fig. \ref{figure.samples} some examples of license plate detection and character detection and recognition using the final proposed system in challenging situations is shown.

\subsection{YOLOv3 results} \label{subsection.YOLOv3results}

The system that is used for training this network is an ASUS notebook with CPU core i7 HQ, GPU GeForce 840M and RAM 8GB. This network trained in Linux operating system using Python and TensorFlow. The proposed YOLOv3 has been trained with 2316 Iranian car images that have been labeled before. The input images have been resized to 320\(\times\)320 pixels at the network entrance. Finally after training, the system have been tasted with 512 Iranian car images and the results have been summarized in table. \ref{table.precision}. 

\begin{table*}[t!]
	\centering
	\renewcommand{\arraystretch}{1.3} 
	\caption{Proposed YOLOv3 test results.}
	\label{table.precision}
	\begin{tabular}{|p{1.1cm}|p{1.5cm}|p{0.4cm}|p{0.4cm}|p{0.4cm}|p{0.95cm}|p{1.05cm}|p{1.2cm}|p{0.6cm}|}
		\hline
		System & Average Detection Speed & TP & FP & FN & Recall & Precision & Average IOU  & mAP  \\
		\hline
		YOLOv3 & 0.23 & 511 & 1 & 9 & 98.26\% & 99.8\% & 81.67\% & 99.6\%\\
		\hline
	\end{tabular}
\end{table*}

\subsection{ Faster R-CNN results} \label{subsection.FasterR-CNNresults}
\begin{figure}[!t]
	\centering
	\includegraphics[width=1\textwidth]{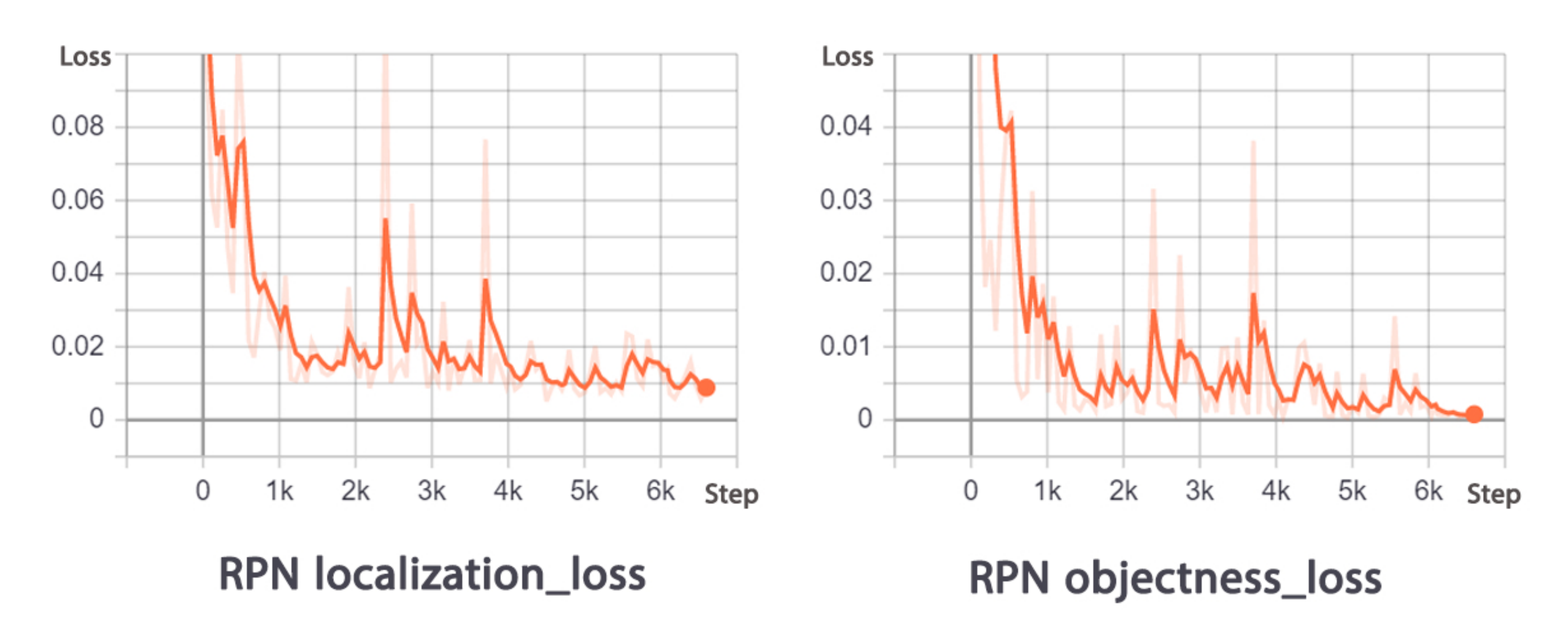}
	\caption{RPN Classification and localization loss. Left plot shows localization loss and right plot shows objectness loss. Pale plots show original losses while bold plots show smoothed losses with smoothing factor equals to 0.6.}
	\label{figure.RPNloss}
\end{figure}

The same ASUS notebook system and same software has been used for training Faster R-CNN network. The proposed Faster R-CNN network trained with 1643 Iranian license plate images that have 13144 objects (numbers and alphabets) that have been labeled before. The input plate images resized to 128\(\times\)640 pixels at the network entrance. Finally after training, the system have been tested with 517 plates and 4103 objects (numbers and alphabets). 
As mentioned before, Faster R-CNN have two networks RPN and Box-Classifier (Fast R-CNN detector), RPN model has two output one is for classification whether it is an object or not and the other one is for bounding box coordinates regression. In Fig. \ref{figure.RPNloss} two chart is shown that one is for objectness loss and the other one is for localization loss. As shown in Fig. \ref{figure.RPNloss}, it learned fast in 1350 epochs, then the learning rate became slower. At final after 6597 epochs the localization loss come down to 0.008567\% and objectness-loss become 0.00077695\%.  

\begin{figure}[h!]
	\centering
	\includegraphics[width=1\textwidth]{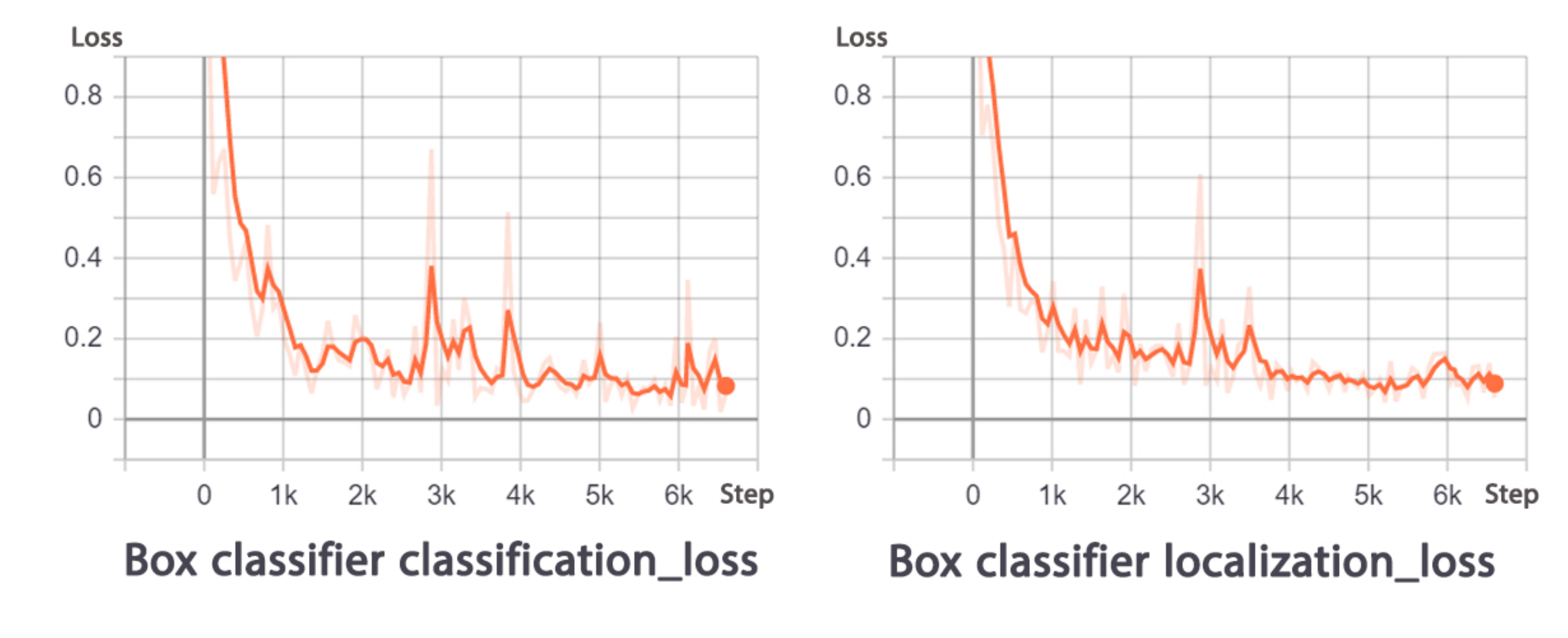}
	\caption{Faster R-CNN object detector (box-classifier) loss. Left plot shows box classification loss and right plot shows localization loss. Pale plots show original losses while bold plots show smoothed losses with smoothing factor equals to 0.6.}
	\label{figure.classifierloss}
\end{figure}

Also Box-Classifier model has two output. One is for classification that determine each box belongs to which class and the other one is for bounding box coordinates regression. In the Fig. \ref{figure.classifierloss}, two charts are shown that one is for classification loss and the other one is localization loss. At final after 6597 epoch the localization loss come down to 0.08873\% and classification-loss reaches 0.08292\%. 

\begin{figure}[h!]
	\centering
	\includegraphics[scale=0.3]{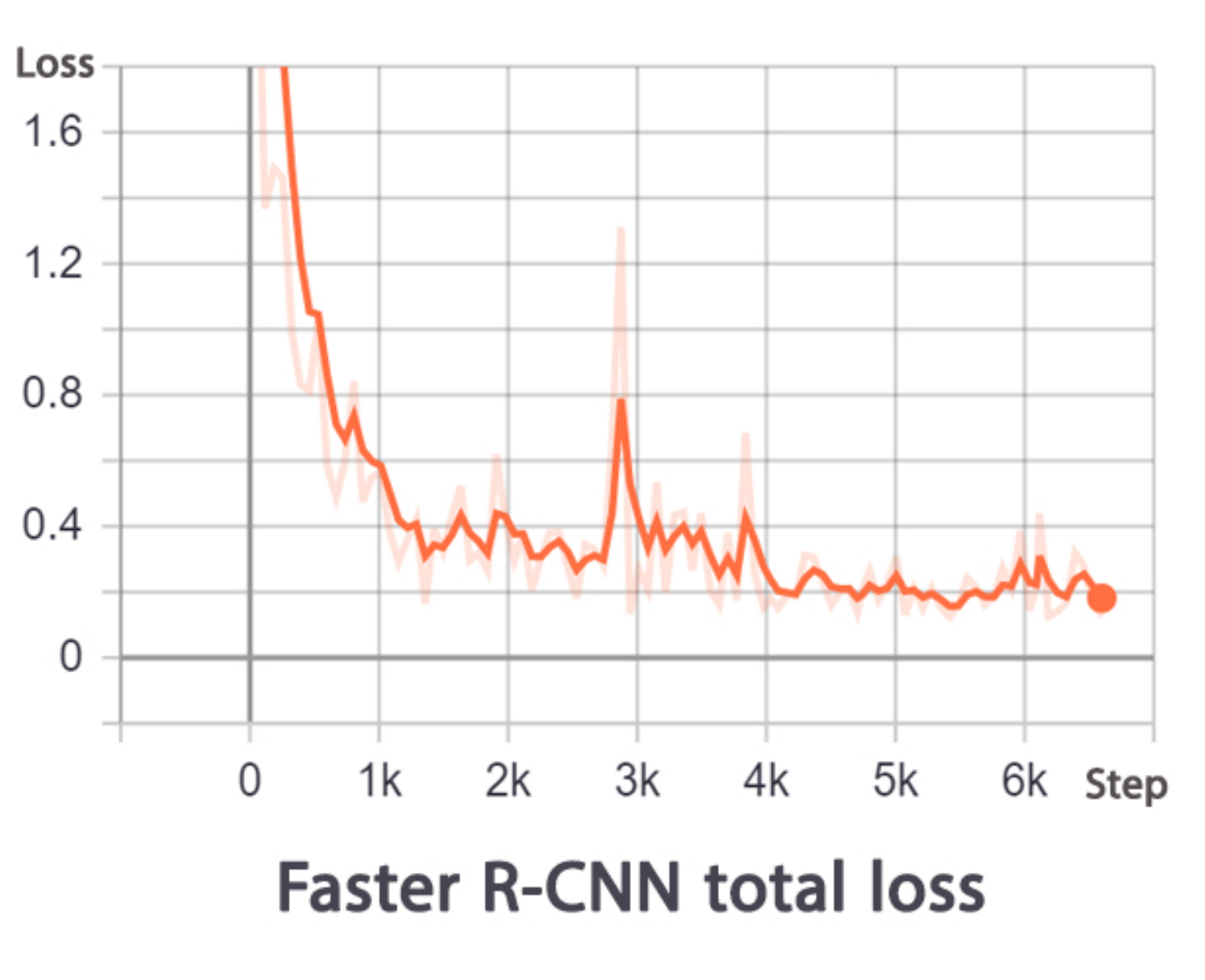}
	\caption{Faster R-CNN total loss. Pale plots show original losses while bold plots show smoothed losses with smoothing factor equals to 0.6. }
	\label{figure.FasterRCNNloss}
\end{figure}

Now in Fig. \ref{figure.FasterRCNNloss} total loss is shown, total loss is sum of RPN losses and box classifier losses. It can be seen that after the 6600 epochs total loss come down to 0.18\%. 

Again after training; the system have been tasted with 517 plates and 4103 objects(numbers and alphabets) so each Faster R-CNN class results and total Faster R-CNN results has been calculated and show in Table. \ref{table.FasterRCNNclassesresults} and Table. \ref{table.FasterRCNNtotalresults}.

\begin{table*}[!t] 
	\renewcommand{\arraystretch}{1.3} 
	\caption{Faster R-CNN classes results}
	\label{table.FasterRCNNclassesresults}
	\centering
	\begin{tabular}{|p{1.4cm}|p{0.57cm}|p{0.57cm}|p{0.57cm}|p{0.57cm}|p{0.57cm}|p{0.57cm}|p{0.57cm}|p{0.57cm}|p{0.57cm}|p{0.57cm}|}
		\hline
		Number or Alphabets & \includegraphics[scale=0.07]{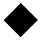} & \includegraphics[scale=0.07]{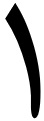} & \includegraphics[scale=0.07]{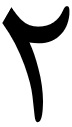} & \includegraphics[scale=0.07]{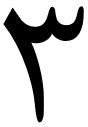} & \includegraphics[scale=0.07]{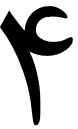} & \includegraphics[scale=0.07]{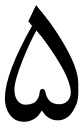} & \includegraphics[scale=0.07]{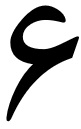} & \includegraphics[scale=0.07]{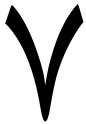} & \includegraphics[scale=0.07]{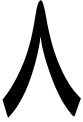} & \includegraphics[scale=0.07]{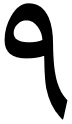}  \\
		\hline
		TP & 134 & 454 & 476 & 373 & 368 & 322 & 330 & 322 & 409 & 367\\
		\hline
		FN & 7 & 2 & 2 & 3 & 3 & 1 & 5 & 2 & 4 & 5\\
		\hline
		FP & 1 & 2 & 0 & 0 & 1 & 0 & 0 & 0 & 0 & 0\\
		\hline
		Acuuracy & 94.3\% & 99.1\% & 99.5\% & 99.2\% & 98.9\% & 99.6\% & 98.5\% & 98.3\% & 99.0\% & 98.6\%\\
		\hline
		Number or Alphabet &\includegraphics[scale=0.07]{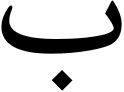} & \includegraphics[scale=0.07]{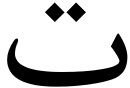}& \includegraphics[scale=0.07]{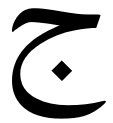} & \includegraphics[scale=0.07]{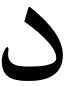} & \includegraphics[scale=0.07]{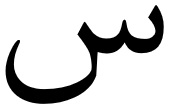} & \includegraphics[scale=0.07]{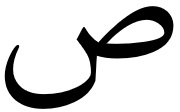} & \includegraphics[scale=0.07]{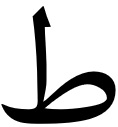} & \includegraphics[scale=0.07]{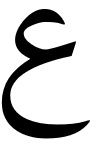} & \includegraphics[scale=0.07]{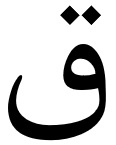} & \includegraphics[scale=0.07]{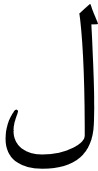}\\
		\hline
		TP &22 &6 &34 &33 & 25 & 63 & 24 & 30 & 46 & 58\\
		\hline
		FN&1 &0 &2 & 0 & 0 & 2 & 0 & 0 & 0 & 0 \\
		\hline
		FP&0 &0 &0 & 0 & 0 & 0 & 0 & 0 & 0 & 0 \\
		\hline
		Acuuracy&95.6\% &100\% &94.4\% & 100\% & 100\% & 96.9\% & 100\% & 100\% &100\% & 100\%\\
		\hline
                Number or Alphabet &\includegraphics[scale=0.07]{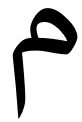} & \includegraphics[scale=0.07]{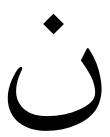} & \includegraphics[scale=0.07]{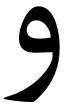} &* &* &* &* &* &* &* \\
               \hline
                TP & 45 & 38 &21 &* & *&* &* &* &* &* \\
               \hline
               FN  & 0& 0& 1& *&* &* &* &* &* &* \\
              \hline
              FP  & 0&0 &0 & *&* &* &* &* &* &* \\
              \hline
               Accuracy  &100\% &100\% &95.4\% &* &* &* &* &* &* &* \\
             \hline

	\end{tabular}
\end{table*}

\begin{table*}[t!]
	\centering
	\renewcommand{\arraystretch}{1.3} 
	\caption{Faster R-CNN totall results }
	\label{table.FasterRCNNtotalresults}
	\begin{tabular}{|p{2cm}|p{0.6cm}|p{0.6cm}|p{0.6cm}|p{1.5cm}|p{1.5cm}|p{1.5cm}|}
		\hline
		System &  TP & FP & FN & Recall & Precision & Acuuracy  \\
		\hline
		Faster R-CNN &  4061 & 4 & 42 & 98.97\% & 99.9\% & 98.86\% \\
		\hline
	\end{tabular}
\end{table*}

\subsection{Comparing proposed system with other systems}
In the subsection\ref{subsection.YOLOv3results},\ref{subsection.FasterR-CNNresults} experiments and results about the YOLOv3 and Faster R-CNN separately explained and the results after training and testing this two networks achieved accuracy of 98.08\% for YOLOv3 network and 98.88\% for Faster R-CNN network.  As mentioned this proposed system made with a YOLOv3 network for detection Iranian license plate from the input image and a Faster R-CNN network for detection numbers and alphabets from cropped license plate and recognition each of them separately. For training YOLOv3 a dataset consist of 2316 images of Iranian car images is used and these images captured in open environment and with different resolutions. The system for training YOLOv3 was an ASUS notebook with CPU core i7 HQ, GPU GeForce 840M and RAM 8GB. After training this system was tasted with 512 Iranian car image. In table.\ref{table.comparedetection} the detection part of the proposed method is compared with some other license plate detection systems. Results show that the proposed YOLOv3 network perform better in comparison with other systems.

\begin{table}[h!]
	\renewcommand{\arraystretch}{1.3} 
	\caption{Comparing the proposed detection system with other systems.}
	\label{table.comparedetection}
	\begin{tabular}{|p{5cm}|p{1.1cm}|p{1.2cm}|}
		\hline
		Method & Accuracy & Reference \\
		\hline
		YOLOv3 & 98.08\% & Proposed \\
		\hline
		Color feature  & 96.8\% & \cite{ref.colorfeatures}\\
		\hline
		Horizontal projection   & 88\% & \cite{ref.templatematching1}\\
		\hline
		Horizontal projection & 97.3\% & \cite{ref.templatematching2}\\
		\hline
		Edge detection with connected component & 91\% & \cite{ref.CNN2}\\
		\hline
		Vertical edge and morphological & 95.2\% & \cite{ref.morphologhy}\\
		\hline
	\end{tabular}
\end{table}

For training Faster R-CNN a dataset consists of 1643 images of Iranian License plate that have been already cropped with YOLOv3 are used. This dataset consists of 13144 objects in total. After training, this system was tested with 517 Iranian license plate images and total loss come down to 0.18\%. In Table. \ref{table.comparerecognition} the proposed method compared with some Iranian license plate recognition systems. As it can be seen from the results, the proposde Faster R-CNN network outperforms other systems.

\begin{table}[h!]
	\renewcommand{\arraystretch}{1.3} 
	\caption{Comparing the proposed recognition system with other systems.}
	\label{table.comparerecognition}
	\begin{tabular}{|p{5cm}|p{1.1cm}|p{1.2cm}|}
		\hline
		Method & Accuracy & Reference \\
		\hline
		Faster R-CNN  & 98.86\% & Proposed \\
		\hline
		Decision tree and SVM  & 94.4\% & \cite{ref.colorfeatures}\\
		\hline
		Template Matching   & 96.02\% & \cite{ref.templatematching1}\\
		\hline
		Template Matching & 93\% & \cite{ref.templatematching2}\\
		\hline
		Hybrid KNN and SVM & 97.03\% & \cite{ref.KNN-SVM}\\
		\hline
		CNN & 97\% & \cite{ref.CNN2}\\
		\hline
	\end{tabular}
	
\end{table}

\subsection{Performance in presence of noise}
In this subsection, we have analyzed the performance of the proposed framework in presence of noise. To this end, Gaussian noise ($SNR=30dB$) is added to the dataset images. Without adding any noise filtering techniques, results show that the proposed framework can still recognize the license plates effectively. Table. \ref{table.noise} shows the comparison results for two cases: noisy and noiseless data.

\begin{table}[h!]
	\renewcommand{\arraystretch}{1.3} 
	\caption{Performance of the proposed framework in presence of Gaussian noise.}
	\label{table.noise}
	\begin{tabular}{|p{2.5cm}|p{1.1cm}|p{1.2cm}|}
		\hline
		Method & Accuracy & Data \\
		\hline
		Faster R-CNN  & 97.24\% & Noisy \\
		\hline
		YOLOv3  & 96.37\% & Noisy\\
		\hline
		Faster R-CNN  & 98.86\% & Noiseless \\
		\hline
		YOLOv3  & 98.08\% & Noiseless\\
		\hline
	\end{tabular}
	
\end{table}

\section{Conclusion} \label{section.conclusion}
In this paper a license plate recognition system based on deep convolutional neural networks is proposed. In first main part a YOLOv3 network has been used for detection of license plate from input image. YOLOv3 uses Darknet53 as a feature extractor. This network made the detection part real time also this system succeeded to catch an accuracy up to 98\%. In the second main part a Faster R-CNN has been used for recognition of the plates that are cropped from the first network. Faster R-CNN uses ResNET101 as a feature extractor. This system also reaches 98.86\% accuracy. Also in this paper two datasets for training the detection and recognition parts of the proposed method has been gathered and labeled. The experimental results show that the proposed system has reached high performance on recognition speed and accuracy in comparison with other Iranian license plate recognition systems, which can fully meet the needs of the practical applications.


%
%

\bibliographystyle{spmpsci}      
\bibliography{Ref}   


\end{document}